\documentclass[11pt,a4paper]{article}
\usepackage[hyperref]{acl2020}
\usepackage{times}
\usepackage{latexsym}

\usepackage{microtype}

\aclfinalcopy % Uncomment this line for the final submission
\def\aclpaperid{834} %  Enter the acl Paper ID here

\usepackage{tikz-dependency}
\usepackage{latexsym}
\usepackage{amssymb}
\usepackage{amsmath}
\mathchardef\mhyphen="2D 
\newcommand{\transname}[1]{\ensuremath{\mathsf{#1}}}
\newcommand{\at}{\transname{Attach}\mhyphen\textit{p}}
\newcommand{\atn}{\transname{Attach}\mhyphen}
\newcommand{\sh}{\transname{Shift}}

\title{Transition-based Semantic Dependency Parsing with Pointer Networks}

\author{Daniel Fern\'{a}ndez-Gonz\'{a}lez \and Carlos G\'{o}mez-Rodr\'{i}guez\\
	Universidade da Coru\~{n}a, CITIC\\
	FASTPARSE Lab, LyS Group \\
Depto. de Ciencias de la Computaci\'{o}n y Tecnolog\'{i}as de la Informaci\'{o}n \\
	Campus de Elvi\~{n}a, s/n, 15071 A Coru\~{n}a, Spain \\
  {\tt d.fgonzalez@udc.es}, {\tt carlos.gomez@udc.es}\\}

\date{}

\begin{document}
\maketitle
\begin{abstract}
  Transition-based parsers implemented with Pointer Networks
  have become the new state of the art in dependency parsing, excelling in producing labelled syntactic trees and outperforming graph-based models in this task. In order to further test the capabilities of these powerful neural networks on a harder NLP problem, we propose a transition system that, thanks to Pointer Networks, 
  can straightforwardly produce labelled directed acyclic graphs and perform semantic dependency parsing. In addition, we enhance our approach with deep contextualized word embeddings extracted from BERT. The resulting system not only outperforms all existing transition-based models, but also 
  matches the best fully-supervised accuracy to date on the SemEval 2015 Task 18 English datasets among previous state-of-the-art graph-based parsers.
\end{abstract}

\section{Introduction}
\begin{figure}
\centering
\small
\includegraphics[width=1.0\columnwidth]{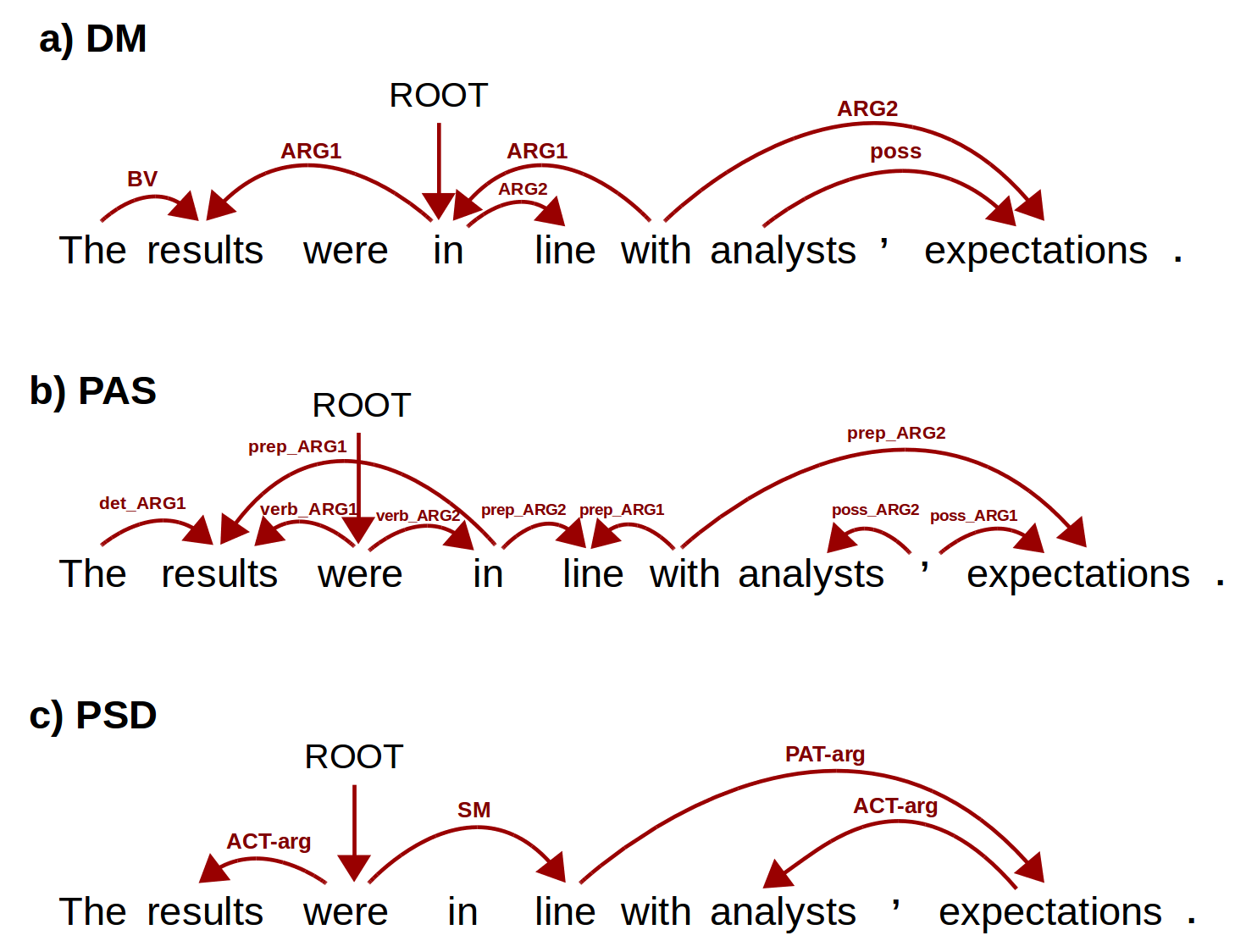}
\caption{Sentence from the SemEval 2015 Task 18 development set parsed with semantic dependencies following the DM, PAS and PSD formalisms.}
\label{fig:sentence}
\end{figure}

In \textit{dependency parsing}, the syntactic structure of a sentence is represented by means of a labelled tree, where each word is forced to be attached exclusively to another that acts as its head. In contrast, \textit{semantic dependency parsing} (SDP) \cite{Oepen2014} aims to represent binary predicate-argument relations between words of a sentence,
which requires producing a labelled directed acyclic graph (DAG):
not only semantic predicates can have multiple or zero arguments, 
but words from the sentence can be attached as arguments to more than one head word (predicate),
or they can be outside the SDP graph (being neither a predicate nor an argument) as shown in the examples in Figure~\ref{fig:sentence}. Since existing dependency parsers cannot be directly applied, 
most SDP research
has focused on adapting them to deal with the absence of single-head and connectedness constraints and to produce an SDP graph instead.

As in dependency parsing, we can find two main families of approaches to efficiently generate accurate SDP graphs. On the one hand, \textit{graph-based} algorithms have drawn more attention since 
adapting them to this task is relatively straightforward.
In particular, these globally optimized methods independently score 
arcs (or sets of them) and then search for a
high-scoring graph
by combining these scores.
From one of the first graph-based DAG parsers proposed by \citet{mcdonald-pereira-2006-online} to the current state-of-the-art models \cite{Wang2019, He2019}, different graph-based SDP approaches have been presented, providing accuracies above their main competitors: \textit{transition-based} DAG algorithms.

A transition-based parser generates a sequence of actions to incrementally build a valid graph (usually from left to right). This is typically done by local, greedy prediction and can efficiently parse a sentence 
in a linear or quadratic number of actions (transitions);
however, the 
lack of global inference makes them more prone to suffer from error propagation: \textit{i.e.}, since transitions are sequentially and locally predicted, 
an erroneous
action can affect future predictions, having a significant impact in long sentences and being, to date, less appealing for SDP. In fact, 
in recent years 
only a few contributions, such as
the system developed by \citet{Wang2018}, present a purely transition-based SDP parser. It is more common to find hybrid systems that combine transition-based approaches with graph-based techniques to alleviate the impact of error propagation in 
accuracy \cite{Du2015}, 
but this penalizes
the efficiency provided by transition-based algorithms.

Away from the current mainstream,
we present a purely transition-based parser that directly generates SDP graphs without the need of any additional techniques. We rely on Pointer Networks \cite{Vinyals15} to predict transitions that can attach multiple heads to the same word and incrementally build a labelled DAG. This kind of neural networks provide an encoder-decoder architecture that is capable of capturing information from
the whole sentence and previously created arcs, alleviating the impact of error propagation and already showing remarkable results in transition-based dependency parsing \cite{Ma18,L2RPointer}.
We 
further enhance our neural network with deep contextualized word embeddings extracted from the pre-trained language model BERT \cite{BERT}.

The proposed SDP parser\footnote{Source code available at \url{https://github.com/danifg/SemanticPointer}.}
can process 
sentences in SDP treebanks (where structures are sparse DAGs with a low in-degree) in $O(n^2 \mathit{log}\ n)$ time, or $O(n^2)$ without cycle detection. This is more efficient than the current fully-supervised
state-of-the-art system by \citet{Wang2019} ($O(n^3)$ without cycle detection), while matching
its accuracy 
on the SemEval 2015 Task 18 datasets \cite{Oepen2015}. In addition, we also prove that our novel transition-based model provides promising accuracies in the semi-supervised scenario, achieving some state-of-the-art results. 

\section{Related Work}
\label{sec:relwork}
An early approach to DAG parsing was implemented as a modification to a graph-based parser by
\citet{mcdonald-pereira-2006-online}. This produced DAGs using
approximate inference by
first finding a dependency tree,
and then adding extra edges that would increase the graph's overall score. A few years later, this 
attempt was outperformed by the first transition-based DAG parser by \citet{sagae-tsujii-2008-shift}. They extended the existing transition system by \citet{nivre-2003-efficient} to allow multiple heads per token. The resulting algorithm was not able to produce DAGs with crossing dependencies, requiring the pseudo-projective transformation by \citet{nivre-nilsson-2005-pseudo} (plus a cycle removal procedure) as a post-processing stage.

More recently, there has been a 
predominance of
purely graph-based DAG models since the SemEval 2015 Task 18 \cite{Oepen2015}. \citet{Almeida2015} 
adapted the pre-deep-learning dependency parser by \citet{martins-etal-2013-turning} to produce SDP graphs. This graph-based parser encodes higher-order information with hand-crafted features and employs the $AD^3 $ algorithm \cite{martins-etal-2011-dual} to find valid DAGs during decoding. This was extended by \citet{Peng2017}
with BiLSTM-based feature extraction and multitask learning: the three formalisms considered in the shared task were jointly learned to improve final accuracy. 

After the success of \citet{dozat-etal-2017-stanfords} in graph-based dependency parsing, \citet{Dozat2018} proposed minor adaptations to use this biaffine neural architecture to produce SDP graphs. To that end, they removed the maximum spanning tree algorithm \cite{chu65,edmonds67} necessary for decoding well-formed dependency trees and simply kept those edges with a positive score. In addition, they trained the unlabelled parser with a sigmoid cross-entropy (instead of the original softmax one) in order to accept multiple heads. 

The parser by \citet{Dozat2018} was recently improved by two contributions. Firstly, \citet{Wang2019} manage to add second-order information for score computation and then apply either mean field variational inference or loopy belief propagation information to decode the highest-scoring SDP graph. While significantly boosting parsing accuracy, the original $O(n^2)$ runtime complexity is modified to $O(n^3)$ in the resulting SDP system. Secondly, \citet{He2019} significantly improve the original parser's accuracy by not only using contextualized word embeddings extracted from BERT \cite{BERT}, but also introducing contextual string embeddings (called Flair) \cite{akbik-etal-2018-contextual}, which consist in a novel type of word vector representations based on character-level language modeling. Both extensions, \cite{Wang2019} and \cite{He2019}, 
are currently the state of the art on the SemEval 2015 Task 18 in the fully-supervised and semi-supervised scenarios, respectively.

\citet{Kurita2019} have also recently proposed a complex approach that iteratively applies the syntactic dependency parser by \citet{Zhang17}, sequentially building a DAG structure. At each iteration, the graph-based parser selects the highest-scoring arcs, keeping the single-head constraint. The process ends when no arcs are added in the last iteration. The combination of partial parses results in an SDP graph. Since the graph is built in a sequential process, they use reinforcement learning to guide the model through more optimal paths. Following \citet{Peng2017}, multi-task learning is also added to boost final accuracy.

On the other hand, the use of transition-based algorithms in the SDP task had been less explored until very recently. \citet{Du2015} presented a voting-based ensemble of fourteen graph- and transition-based parsers. In their work, they noticed that individual graph-based models outperform transition-based algorithms, assigning, during voting, higher weights to them. Among the transition systems used, we can find the one developed by \citet{titov09}, which is not able to cover all SDP graphs.

We have to wait until the work by \citet{Wang2018}  to see that a purely transition-based SDP parser (enhanced with a simple model ensemble technique) can achieve competitive results. They simply modified the preconditions of the complex transition system by \citet{Choi2013} to produce unrestricted DAG structures. In addition, their system was implemented by means of stack-LSTMs \cite{dyer-etal-2015-transition}, enhanced with BiLSTMs and Tree-LSTMs for feature extraction. 

We are, to the best of our knowledge, first to explore DAG parsing with Pointer Networks, proposing a purely transition-based algorithm that can be a competitive alternative to graph-based SDP models.

Finally, during the reviewing process of this work, the proceedings of the CoNLL 2019 shared task \cite{oepen-etal-2019-mrp} were released. In that event, SDP parsers were evaluated on updated versions of SemEval 2015 Task 18 datasets, as well as on datasets in other semantic formalisms such as Abstract Meaning Representation (AMR) \cite{AMR} and Universal Cognitive Conceptual Annotation (UCCA) \cite{UCCA}. Although graph-based parsers achieved better accuracy in the SDP track, several BERT-enhanced transition-based  approaches were proposed. Among them we can find an extension \cite{che-etal-2019-hit} of the system by \citet{Wang2018}, several adaptations for SDP \cite{hershcovich-arviv-2019-tupa,bai-zhao-2019-sjtu} of the transition-based UCCA parser by \citet{TUPA}, as well as an SDP variant \cite{lai-etal-2019-cuhk} of the constituent transition system introduced by \citet{nonbinary}. Also in parallel to the development of this research, \citet{transducer} proposed a transition-based parser that, while it can be applied for SDP, was specifically designed for AMR and UCCA parsing (where graph nodes do not correspond with words and must be generated during the parsing process). In particular, this approach incrementally builds a graph by predicting at each step a semantic relation composed of the target and source nodes plus the arc label. While this can be seen as an extension of our approach for those tasks where nodes must be generated, its complexity penalizes accuracy in the SDP task.

\section{Multi-head Transition System}
We design a novel transition system that is able to straightforwardly attach multiple heads to each word in a single pass, incrementally building, from left to right, a valid SDP graph: a labelled DAG. 

To implement it, we use Pointer Networks \cite{Vinyals15}. These neural networks are able to learn the conditional probability of a sequence of discrete numbers that correspond to positions in an input sequence and, at decoding time, perform as a pointer that selects a position from the input. In other words, we can train this neural network to, given a word, point to the position of the sentence where its head \cite{L2RPointer} or dependent words \cite{Ma18} are located, depending on what interpretation we use during training. In particular, \cite{L2RPointer} proved to be more suitable for dependency parsing than \cite{Ma18} since it requires half as many steps to produce the same dependency parse, being not only faster, but also more accurate (as this mitigates the impact of error propagation).

Inspired by \citet{L2RPointer}, we train a Pointer Network to point to the head of a given word and propose an algorithm that does not use any kind of data structures (stack or buffer, required in classic transition-based parsers \cite{nivre08cl}), but just a \textit{focus word pointer i} for marking the word currently being processed. More in detail, given an input sentence of $n$ words $w_1, \dots ,w_n$, the parsing process starts with $i$ pointing at the first word $w_1$. At each time step,
the current focus word $w_i$ is used by the Pointer Network to return a position $p$ from the input sentence (or 0, where the ROOT node is located). This information is used to choose between the two available transitions:
\begin{itemize}
    \item If $p \neq i$,
    then the pointed word $w_p$ is considered as a semantic head word (predicate) of $w_i$ and an $\at$ transition is applied, creating the directed arc $w_p \rightarrow w_i$. The $\at$ transition is only permissible if the resulting predicate-argument arc neither exists nor generates a cycle in the already-built graph, in order to output a valid DAG.

    \item On the contrary,  
    if $p = i$ (\textit{i.e.}, the model points to the current focus word), 
    then 
    $w_i$ is considered to have found all its head words, and a $\sh$ transition is chosen to move $i$ one position to the right to process the next word $w_{i+1}$.
\end{itemize}
\noindent The parsing ends when the last word from the sentence is shifted, meaning that the input is completely processed. As stated by \citet{Ma18} for attaching dependent words, it is necessary to fix the order in which (in our case, head) words are assigned in order to define a deterministic decoding. As the sentence is parsed in a left-to-right manner, we adopt the same order for head assignments. For instance, the SDP graph in Figure~\ref{fig:sentence}(a) is produced by the transition sequence described in Table~\ref{tab:seq}. We just need $n$ $\sh$ transitions to move the focus word pointer through the whole sentence and $m$ $\at$ transitions to create the $m$ arcs present in the SDP graph.

\begin{table}
\begin{center}
\small
\begin{tabular}{@{\hskip 1.0pt}lcc@{\hskip 5.5pt}c@{\hskip 0pt}}
\hline\noalign{\smallskip}
$p$ & transition & focus word$_i$  & added arc \\
\noalign{\smallskip}\hline\noalign{\smallskip}
 & & The$_1$ &  \\
1 & $\sh$ & results$_2$ & \\
1 & $\atn{1}$ & results$_2$ &  The$_1$ $\rightarrow$ results$_2$ \\
4 & $\atn{4}$ & results$_2$ & results$_2$ $\leftarrow$ in$_4$ \\
2 & $\sh$ & were$_3$ & \\
3 & $\sh$ & in$_4$ & \\
0 & $\atn{0}$ & in$_4$ & ROOT$_0$ $\rightarrow$ in$_4$\\
6 & $\atn${6} & in$_4$ & in$_4$ $\leftarrow$ with$_6$\\
4 & $\sh$ & line$_5$ & \\
4 & $\atn{4}$ & line$_5$ & in$_4$ $\rightarrow$ line$_5$\\
5 & $\sh$ & with$_6$ & \\
6 & $\sh$ & analysts$_7$ & \\
7 & $\sh$ & '$_8$ & \\
8 & $\sh$ & expectations$_9$ & \\
6 & $\atn{6}$ & expectations$_9$ & with$_6$ $\rightarrow$ expectations$_9$\\
7 & $\atn{7}$ & expectations$_9$ & analysts$_7$ $\rightarrow$ expectations$_9$\\
9 & $\sh$ & .$_{10}$ & \\
10 & $\sh$ &  & \\
\noalign{\smallskip}\hline
\end{tabular}
\caption{Transition sequence for generating the SDP graph in Figure~\ref{fig:sentence}(a).} \label{tab:seq}
\end{center}
\end{table}

It is worth mentioning that we manage to significantly reduce the amount of transitions necessary for generating DAGs
in comparison to those proposed in the complex transition systems by \citet{Choi2013} and \citet{titov09}, used in the SDP systems by \citet{Wang2018} and \citet{Du2015}, respectively. In addition, the described multi-head transition system is able to directly produce any DAG structure without exception, while some transition systems, such as the mentioned \cite{sagae-tsujii-2008-shift,titov09}, are limited to a subset of DAGs.

Finally, while the outcome of the proposed transition system is a SDP graph without cycles, in other research, such as \cite{Kurita2019} and state-of-the-art models by \citet{Dozat2018} and \citet{Wang2019}, the parser is not forced to produce well-formed DAGs, allowing the presence of cycles.

\section{Neural Network Architecture}
\subsection{Basic Approach}
\citet{Vinyals15} introduced an encoder-decoder architecture, called \textit{Pointer Network}, that uses a mechanism of neural attention \cite{Bahdanau2014} to select positions from the input sequence, without requiring a fixed size of the output dictionary. This allows Pointer Networks to easily address those problems where the target classes considered at each step are variable and depend on the length of the input sequence. We prove that implementing the transition system previously defined on this neural network results in an accurate SDP system.

We follow previous work in dependency parsing \cite{Ma18,L2RPointer} to design our neural architecture:
\paragraph{Encoder} A BiLSTM-CNN architecture \cite{Ma2016} is used to encode the input sentence $w_1, \dots ,w_n$, word by word, into a sequence of \textit{encoder hidden states} $\mathbf{h}_1, \dots, \mathbf{h}_n$. CNNs with max pooling are used for extracting character-level representations of words and, then, each word $w_i$ is represented by the concatenation of character ($\mathbf{e}^c_i$), word ($\mathbf{e}^w_i$), lemma ($\mathbf{e}^l_i$) and POS tag ($\mathbf{e}^p_i$) embeddings: 
$$\mathbf{x}_i = \mathbf{e}^c_i \oplus \mathbf{e}^w_i \oplus \mathbf{e}^l_i \oplus \mathbf{e}^p_i$$
After that, the $\mathbf{x}_i$ of each word $w_i$ is fed one-by-one into a BiLSTM that captures context information in both directions and generates a vector representation $\mathbf{h}_i$:
$$ \mathbf{h}_i = \mathbf{h}_{li}\oplus\mathbf{h}_{ri} = \mathbf{BiLSTM}(\mathbf{x}_i)$$
In addition, a  special  vector  representation $\mathbf{h}_0$, denoting the ROOT node, is prepended at the beginning of the sequence of encoder hidden states.

\paragraph{Decoder} An LSTM is used to output, at each time step $t$, a \textit{decoder hidden state} $\mathbf{s}_t$. As input of the decoder, we use the encoder hidden state $\mathbf{h}_i$ of the current focus word $w_i$ plus extra high-order features. In particular, we take into account the hidden state of the last head word ($\mathbf{h}_{h}$) attached to $w_i$, which will be a co-parent of a future predicate assigned to $w_i$. Following \citet{Ma18}, we use element-wise sum to add this information without increasing the dimensionality of the input:
$$\mathbf{r}_i = \mathbf{h}_{i} + \mathbf{h}_{h}; \ \mathbf{s}_t = \mathbf{LSTM}(\mathbf{r}_i)$$
Note that feature information like this can be easily added in transition-based models without increasing the parser's runtime complexity, something that does not happen in graph-based models, where, for instance, the second-order features added by \citet{Wang2019} penalize runtime complexity.

We experimented with other high-order features such as grandparent or sibling information of the current focus word $w_i$, but no significant improvements were obtained from their addition, so they were discarded for simplicity. Further feature exploration might improve parser performance, but we leave this for future work.

Once $\mathbf{s}_t$ is generated, the \textit{attention vector} $\mathbf{a}^t$, which will work as a pointer over the input, must be computed in the \textit{pointer layer}. First, following the previously cited work, the scores between $\mathbf{s}_t$ and each encoder hidden representation $\mathbf{h}_j$ from the input sentence
are computed using this \textit{biaffine attention} scoring function \cite{DozatM17}:
\begin{align*}
\mathbf{v}^t_j = \mathbf{score}(\mathbf{s}_t, \mathbf{h}_j)= f_1(\mathbf{s}_t)^T W f_2(\mathbf{h}_j)\\
+\mathbf{U}^Tf_1(\mathbf{s}_t) + \mathbf{V}^Tf_2(\mathbf{h}_j) + \mathbf{b}
\end{align*}
where parameter $W$ is the weight matrix of the bi-linear term, $\mathbf{U}$ and $\mathbf{V}$ are the weight tensors of the linear terms and $\mathbf{b}$ is the bias vector. In addition, 
$f_1(\cdot)$ and $f_2(\cdot)$ are two single-layer multilayer perceptrons (MLP) with ELU activation, proposed by \cite{DozatM17} for reducing dimensionality and minimizing overfitting.

\begin{figure*}[t]
\centering
\includegraphics[width=1.0\textwidth]{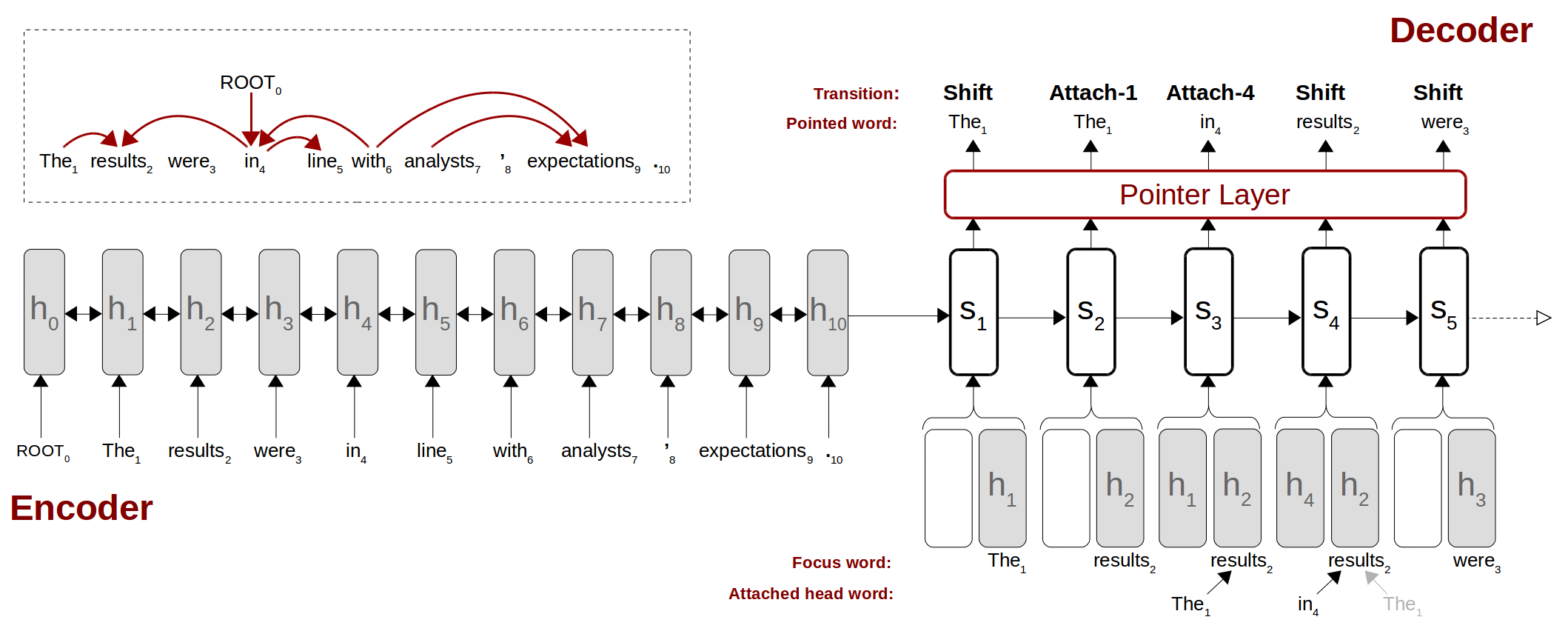}
\caption{Neural network architecture and decoding steps to partially parse the SDP graph in Figure~\ref{fig:sentence}.}
\label{fig:network}
\end{figure*}

Then, a softmax is applied on the resulting score vector $\mathbf{v}^t$ to compute a probability distribution over the input words:
$$\mathbf{a}^t = \mathbf{softmax}(\mathbf{v}^t)$$
The resulting attention vector $\mathbf{a}^t$ can now be used as a pointer to select the highest-scoring position $p$ from the input. This information will be employed by the transition system to choose between the two available actions and create a predicate-argument relation between $w_p$ and $w_i$ ($\at$) or move the focus word pointer to $w_{i+1}$ ($\sh$). In case the chosen $\at$ is forbidden due to the acyclicity constraint, the next highest-scoring position in $\mathbf{a}^t$ is considered as output instead. Figure~\ref{fig:network} depicts the neural architecture and the decoding procedure for the SDP structure in Figure~\ref{fig:sentence}(a).

\paragraph{Label prediction} We jointly train a multi-class classifier that scores every label for each pair of words. This shares the same encoder and uses the same biaffine attention function as the pointer:
\begin{align*}
\mathbf{s}^l_{tp} = \mathbf{score}(\mathbf{s}_t, \mathbf{h}_p, l)= g_1(\mathbf{s}_t)^T W_l g_2(\mathbf{h}_p)\\
+\mathbf{U}_l^Tg_1(\mathbf{s}_t) + \mathbf{V}_l^Tg_2(\mathbf{h}_p) + \mathbf{b}_l
\end{align*} 
where a distinct weight matrix $W_l$, weight tensors $\mathbf{U}_l$ and $\mathbf{V}_l$ and bias $\mathbf{b}_l$ are used for each label $l$, 
where $l \in \{1, 2, \dots , L\}$ and $L$ is the number of labels. 
In addition, 
$g_1(\cdot)$ and $g_2(\cdot)$ are two single-layer MLPs with ELU activation.

The scoring function is applied over each predicted arc between the dependent word $w_i$ (represented by $\mathbf{s}_t$) and the pointed head word $w_p$ in position $p$ (represented by $\mathbf{h}_p$) to compute the score of each possible label and assign the highest-scoring one.

\paragraph{Training Objectives} 
The Pointer Network is trained to
minimize the negative log likelihood 
(implemented as cross-entropy loss) 
of producing the correct SDP graph $y$ for a given sentence $x$: $P_\theta (y|x)$. Let $y$ be a DAG for an input sentence $x$
that is decomposed into a set of $m$ directed arcs $a_1, \dots , a_{m}$ following a left-to-right order.
This probability can be factorized as follows:
$$P_\theta (y|x) = \prod_{k=1}^{m} P_\theta (a_k | a_{<k}, x)$$
where $a_{<k}$ denotes previous predicted arcs. 

On the other hand, the labeler is trained with softmax
cross-entropy to minimize the negative log likelihood of assigning the correct label $l$, given a dependency arc 
with head word $w_h$ and dependent word $w_i$. 

The whole neural model is jointly trained by summing the parser and labeler losses prior to computing the gradients. In that way, model parameters are learned to minimize the sum of the cross-entropy loss objectives over the whole corpus.

\subsection{Deep Contextualized Word Embeddings Augmentation}
In order to further improve the accuracy of our approach, we augment our model with deep contextualized word embeddings provided by the widely-used pre-trained language model BERT \cite{BERT}.

Instead of including and training the whole BERT model as encoder of our system, we follow the common, greener and more cost-effective approach of leveraging the potential of BERT by extracting the weights of one or several layers as word-level embeddings. To that end, the pre-trained uncased BERT$_{\tt BASE}$ model is used. 

Since BERT is trained on \textit{subwords} (\textit{i.e.}, substrings of the original token), we take the 768-dimension vector of each subword of an input token and use the average embedding as the final representation $\mathbf{e}^{BERT}_i$. Finally, this is directly concatenated to the resulting basic word representation before feeding the BiLSTM-based encoder:
$$\mathbf{x}'_i = \mathbf{x}_i \oplus \mathbf{e}^{BERT}_i;\ \mathbf{h}_i = \mathbf{BiLSTM}(\mathbf{x}'_i)$$
Higher performances can be achieved by summing or concatenating (depending on the task) several layers of BERT; however, exploring these combinations is out of the scope of this paper and we simply use embeddings extracted from the second-to-last hidden layer (since the last layer is biased to the target objectives used to train BERT's language model).  

\section{Experiments}
\subsection{Data}
In order to test the proposed approach, we conduct experiments on the SemEval 2015 Task 18 English datasets \cite{Oepen2015}, where all sentences are annotated with three different formalisms: DELPH-IN MRS (DM) \cite{Flickinger2012}, Predicate-Argument Structure (PAS) \cite{Miyao2004} and Prague Semantic Dependencies (PSD) \cite{Hajic2012}. Standard split as in previous work \cite{Almeida2015,Du2015} results in 33,964 training sentences from Sections 00-19 of the Wall Street Journal corpus \cite{marcus93}, 1,692 development sentences from Section 20, 1,410 sentences from Section 21 as in-domain test set, and 1,849 sentences sampled from the Brown Corpus \cite{BrownCorpus} as out-of-domain test data.
For the evaluation, we use the official script,\footnote{\url{https://github.com/semantic-dependency-parsing/toolkit}} reporting labelled F-measure scores (LF1) (including ROOT arcs) on the in-domain (ID) and out-of-domain (OOD) test sets for each formalism as well as the macro-average over the three of them.

\subsection{Settings}
We use the Adam optimizer \cite{Adam} and follow \cite{Ma18,DozatM17} for parameter optimization. We do not specifically perform hyper-parameter selection for SDP and just adopt those proposed by \citet{Ma18} for syntactic dependency parsing (detailed in Table~\ref{tab:hyper}). For initializing word and lemma vectors, we use the pre-trained structured-skipgram embeddings developed by \citet{Ling2015}. POS tag and character embeddings are randomly initialized and all embeddings (except the deep contextualized ones) are fine-tuned during training. Due to random initializations, we report average accuracy over 5 repetitions for each experiment. In addition, during a 500-epoch training, the model with the  highest labelled F-score on the development set is chosen. Finally, while further beam-size exploration might 
improve
accuracy, we use beam-search decoding with beam size 5 in all experiments. 

\begin{table}
\begin{footnotesize}
\centering
\begin{tabular}{@{\hskip 0pt}lc@{\hskip 0pt}}
\hline
\textbf{Architecture hyper-parameters} & \\
\hline
CNN window size & 3 \\
CNN number of filters & 50 \\
BiLSTM encoder layers & 3 \\
BiLSTM encoder size & 512 \\
LSTM decoder layers & 1 \\ 
LSTM decoder size & 512 \\
LSTM layers dropout & 0.33 \\
Word/POS/Char./Lemma embedding dimension & 100\\
BERT embedding dimension & 768\\
Embeddings dropout & 0.33 \\
MLP layers & 1 \\
MLP activation function & ELU \\
Arc MLP size & 512 \\ 
Label MLP size & 128 \\
UNK replacement probability & 0.5 \\
\hline
\textbf{Adam optimizer hyper-parameters} &\\
\hline
Initial learning rate & 0.001 \\
$\beta_1$, $\beta_2$ & 0.9 \\
Batch size & 32 \\
Decay rate & 0.75 \\
Gradient clipping & 5.0 \\
\hline
\multicolumn{1}{c}{}\\
\end{tabular}
\centering
\setlength{\abovecaptionskip}{4pt}
\caption{Model hyper-parameters.}
\label{tab:hyper}
\end{footnotesize}
\end{table}

\subsection{Results and Discussion}
Table~\ref{tab:results} reports the accuracy 
obtained by state-of-the-art SDP parsers detailed in Section~\ref{sec:relwork} in comparison to our approach. To perform a fair comparison, we group SDP systems in three blocks dependending on the embeddings provided to the architecture: (1) just basic pre-trained word and POS tag embeddings, (2) character and pre-trained lemma embeddings augmentation and (3) pre-trained deep contextualized embeddings augmentation.
As proved by these results, our approach outperforms all existing transition-based models and the widely-used approach by \citet{Dozat2018} with or without character and lemma embeddings, and it is on par with the best graph-based SDP parser by \cite{Wang2019} on average in the fully-supervised scenario.\footnote{It is common practice in the literature that systems that only use standard pre-trained word or lemma embeddings are classed as fully-supervised models, even though, strictly, they are not trained exclusively on the official training data.}

In addition, our model achieves the best fully-supervised accuracy to date on the PSD formalism, considered the hardest to parse. We hypothesize that this might be explained by the fact that the PSD formalism is the more tree-oriented (as pointed out by \citet{Oepen2015}) and presents a lower ratio of arcs per sentence, being more suitable for our transition-based approach. 

In the semi-supervised scenario, BERT-based embeddings proved to be more beneficial for the out-of-domain data. In fact, while not being a fair comparison since we neither include contextual string embeddings (Flair) \cite{akbik-etal-2018-contextual} nor explore different BERT layer combinations, our 
new
transition-based parser manages to outperform the state-of-the-art system by \citet{He2019}\footnote{\citet{He2019} do not specify 
in their paper
the BERT layer configuration used for generating the word embeddings.} on average on the out-of-domain test set, obtaining a remarkable accuracy on the PSD formalism.

\begin{table*}
\centering
\begin{tabular}{@{\hskip 0.5pt}lcccccccc@{\hskip 0.5pt}}
& \multicolumn{2}{c}{DM}
& \multicolumn{2}{c}{PAS}
& \multicolumn{2}{c}{PSD}
& \multicolumn{2}{c}{Avg}
\\
Parser & ID & OOD & ID & OOD & ID & OOD & ID & OOD \\
\hline
\citet{Du2015} \scriptsize{TbGb+Ens}  & 89.1 & 81.8 & 91.3 & 87.2 & 75.7 & 73.3 & 85.3 & 80.8 \\
\citet{Almeida2015} \scriptsize{Gb}    & 88.2 & 81.8 & 90.9 & 86.9 & 76.4 & 74.8 & 85.2 & 81.2 \\
\citet{Peng2017} \scriptsize{Gb}   & 89.4 & 84.5 & 92.2 & 88.3 & 77.6 & 75.3 & 86.4 & 82.7 \\
\citet{Peng2017} \scriptsize{Gb+MT}  & 90.4 & 85.3 & 92.7 & 89.0 & 78.5 & 76.4 & 87.2 & 83.6 \\
\citet{Wang2018} \scriptsize{Tb}  & 89.3 & 83.2 & 91.4 & 87.2 & 76.1 & 73.2 & 85.6 & 81.2 \\
\citet{Wang2018} \scriptsize{Tb+Ens}  & 90.3 & 84.9 & 91.7 & 87.6 & 78.6 & 75.9 & 86.9 & 82.8 \\
\citet{Dozat2018} \scriptsize{Gb}   & 91.4 & 86.9 & 93.9 & 90.8 & 79.1 & 77.5 & 88.1 & 85.0 \\
\citet{Kurita2019} \scriptsize{Gb}   & 91.1 & - & 92.4 & - & 78.6 & - & 87.4 & -  \\
\citet{Kurita2019} \scriptsize{Gb+MT+RL}   & 91.2 & - & 92.9 & - & 78.8 & - & 87.6 & - \\
\citet{Wang2019} \scriptsize{Gb}   & \textbf{93.0} & \textbf{88.4} & \textbf{94.3} & \textbf{91.5} & 80.9 & \textbf{78.9} & \textbf{89.4} & \textbf{86.3} \\
\textbf{This work} \scriptsize{Tb}   & 92.5 & 87.7 & 94.2 & 91.0 & \textbf{81.0} & 78.7  & 89.2 & 85.8 \\
\hline
\citet{Dozat2018} \scriptsize{Gb\textbf{+char+lemma}}   & 93.7 & 88.9 & 93.9 & 90.6 & 81.0 & 79.4 & 89.5 & 86.3 \\
\citet{Kurita2019} \scriptsize{Gb+MT+RL\textbf{+lemma}}   & 92.0 & 87.2 & 92.8 & 88.8 & 79.3 & 77.7 & 88.0 & 84.6 \\
\citet{Wang2019} \scriptsize{Gb\textbf{+char+lemma}}   & \textbf{94.0} & \textbf{89.7} & 94.1 & \textbf{91.3} & 81.4 & 79.6 & 89.8 & \textbf{86.9} \\
\textbf{This work} \scriptsize{Tb\textbf{+char+lemma}}    & 93.9 & 89.6 & \textbf{94.2} & 91.2 & \textbf{81.8} & \textbf{79.8} & \textbf{90.0} & \textbf{86.9} \\
\hline
\citet{transducer} \scriptsize{Tb+char\textbf{+BERT$_{\tt LARGE}$}}   & 92.2 & 87.1 & - & - & - & - & - & - \\
\citet{He2019} \scriptsize{Gb+lemma\textbf{+Flair+BERT$_{\tt BASE}$}}   & \textbf{94.6} & 90.8 & \textbf{96.1} & \textbf{94.4} & \textbf{86.8} & 79.5 & \textbf{92.5} & 88.2 \\
\textbf{This work} \scriptsize{Tb+char+lemma\textbf{+BERT$_{\tt BASE}$}}   & 94.4  & \textbf{91.0} & 95.1 & 93.4 & 82.6 & \textbf{82.0} & 90.7 & \textbf{88.8} \\
\hline
\multicolumn{1}{c}{}\\
\end{tabular}
\centering
\setlength{\abovecaptionskip}{4pt}
\caption{Accuracy comparison of state-of-the-art SDP parsers on the SemEval 2015 Task 18 datasets. $Gb$ and $Tb$ stand for graph- and transition-based models, $+\mathit{char}$ and $+\mathit{lemma}$ for augmentations with character-level and lemma embeddings, $+\mathit{Flair}$ and $+\mathit{BERT}_{\tt BASE|LARGE}$ for augmentations with deep contextualized character-level and word-level embeddings, and, finally, $+\mathit{MT}$, $+\mathit{RL}$ and $+\mathit{Ens}$ for the application of multi-task, reinforcement learning and ensemble techniques.
}
\label{tab:results}
\end{table*}

\subsection{Complexity}

\begin{figure}
\centering
\includegraphics[width=0.45\textwidth]{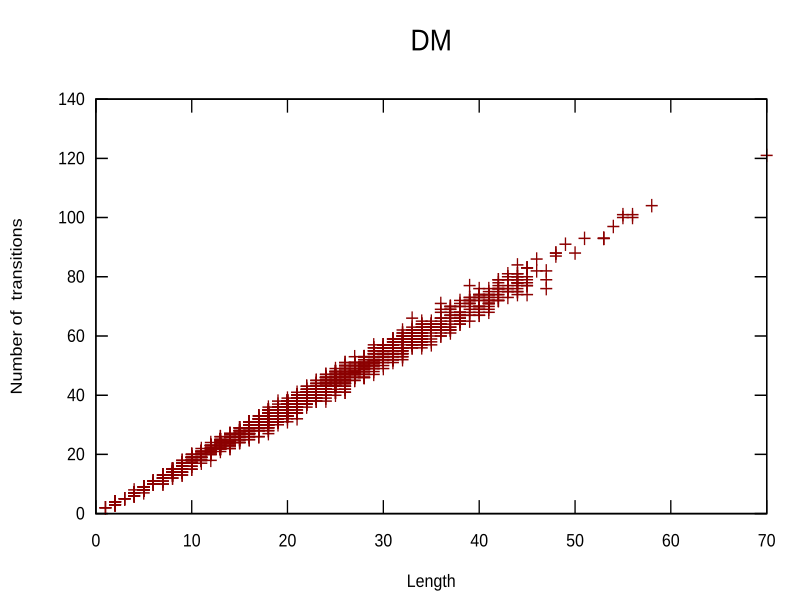}
\includegraphics[width=0.45\textwidth]{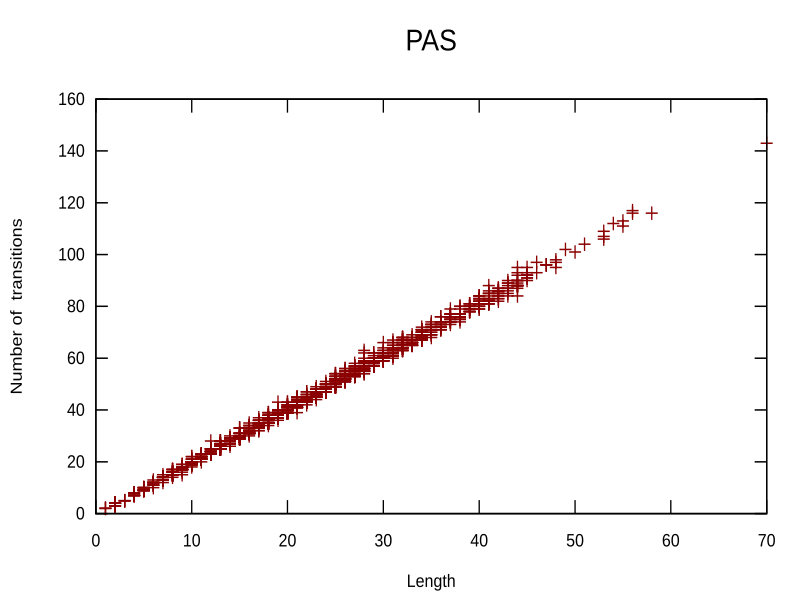}
\includegraphics[width=0.45\textwidth]{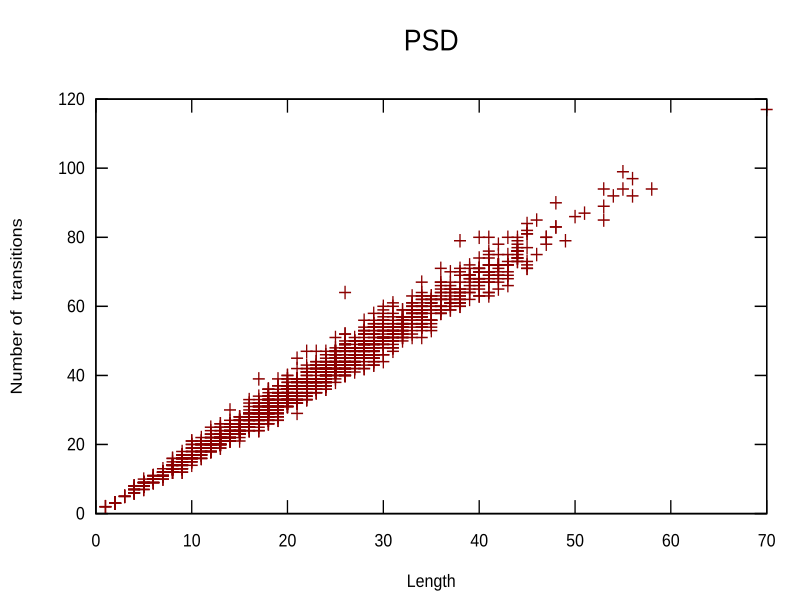}
\caption{Number of predicted transitions relative to the length of the sentence, for the three SDP formalisms on the development set from SemEval 2015 Task 18.}
\label{fig:run}
\end{figure}

Given a sentence with length $n$ whose SDP graph has $m$ arcs, the proposed transition system requires $n$ $\sh$ plus $m$ $\at$ transitions to parse it. 
Therefore, since a DAG can have at most $\Theta(n^2)$ edges (as is also the case for general directed graphs), it could potentially need $O(n^2)$ transitions in the worst case. However, we prove that this does not happen in practice and real sentences can be parsed with $O(n)$ transitions instead.

Parsing complexity of a transition-based dependency parsing algorithm can be determined by the number of transitions performed with respect to the number of words in a sentence \cite{Kubler2009}. Therefore, we measure the transition sequence length predicted by the system to analyze every sentence from the development sets of the three available formalisms and depict the relation between them and sentence lengths. As shown in Figure~\ref{fig:run}, a linear behavior is observed in all cases, proving that the number of $\at$ transitions evaluated by the model at each step is considerably low (behaving practically like a constant). 

This can be explained by the fact that, on average on the training set, the ratio of predicate-argument dependencies per word in a sentence is 0.79 in DM, 0.99 in PAS and 0.70 in PSD, meaning that the transition sequence necessary for parsing a given sentence will need no more
$\at$ transitions than $\sh$ ones (which are 
one per word in the sentence).
It is true that one argument can be attached to more than one predicate; however, the amount of words unattached 
in the resulting DAG (singletons)\footnote{A singleton is a word that has neither incoming nor outgoing edges.} can be significant in some formalisms
(as described graphically in Figure~\ref{fig:sentence}): on average on the training set, 23\% of words per sentence in DM, 6\% in PAS and 35\% in PSD. In addition, edge density on non-singleton words, computed by \citet{Oepen2015} on the test sets, also backs the linear behavior shown in our experiments: 0.96 in DM, 1.02 in PAS and 1.01 in PSD for the in-domain set and 0.95 in DM, 1.02 in PAS and 0.99 in PSD for the out-of-domain data. In conclusion, we can state that, on the datasets tested, the proposed transition system 
executes $O(n)$ transitions.

To determine the runtime complexity of the implementation of the transition system, we need to consider the following: firstly, at each transition, the attention vector $\mathbf{a}^t$ needs to be computed, which means that each of the $O(n)$ transitions takes $O(n)$ time to run. Therefore, the overall time complexity of the parser, ignoring cycle detection, is $O(n^2)$. Note that this is in contrast to algorithms like \cite{Wang2019}, which takes cubic time even though it does not enforce acyclicity.

If we add cycle detection, needed to forbid transitions that would create cycles and therefore to enforce that the output is a DAG, then the complexity becomes $O(n^2 \mathit{log}\ n)$. This is because an efficient implementation of cycle detection contributes an additive factor of $O(n^2 \mathit{log}\ n)$ to worst-case time complexity, which becomes the dominant factor. To achieve this efficient implementation, we incrementally keep two data structures: on the one hand, we keep track of weakly connected components using path compression and union by rank, which can be done in inverse Ackermann time, as is commonly done for cycle detection in tree and forest parsers \cite{covington01,gomez-rodriguez-nivre-2010-transition}. On the other hand, we keep a weak topological numbering of the graph using the algorithm by \citet{Bender2015}, which takes overall $O(n^2 \mathit{log}\ n)$ time over all edge insertions. When these two data structures are kept, cycles can be checked in constant time: an arc $a \rightarrow b$ creates a cycle if the involved nodes are in the same weakly connected component and $a$ has a greater topological number than $b$.

Therefore, the overall expected worst-case running time of the proposed SDP system is $O(n^2 \mathit{log}\ n)$ for the range of data attested in the experiments, and can be lowered to $O(n^2)$ if we are willing to forgo enforcing acyclicity.

\section{Conclusions and Future work}
Our multi-head transition system can accurately parse a sentence in quadratic worst-case runtime
thanks to Pointer Networks. While being more efficient, our approach outperforms the previous state-of-the-art parser by \citet{Dozat2018} and matches the accuracy of the best model to date \cite{Wang2019}, proving that, with a state-of-the-art neural architecture, transition-based SDP parsers are a competitive alternative.

By adding BERT-based embeddings, we significantly improve our model accuracy by marginally affecting computational cost, achieving state-of-the-art F-scores in out-of-domain test sets.

Despite the promising results, the accuracy of our approach 
could probably be boosted further
by experimenting with new feature information and specifically tuning hyper-parameters for the SDP task, as well as using different enhancements such as implementing the hierarchical decoding recently presented by \citet{Liu2019}, including contextual string embeddings \cite{akbik-etal-2018-contextual} like \citet{He2019}, or applying multi-task learning across the three formalisms like
\citet{Peng2017}.

\section*{Acknowledgments}
This work has received funding from the European
Research Council (ERC), under the European
Union's Horizon 2020 research and innovation
programme (FASTPARSE, grant agreement No
714150), from the ANSWER-ASAP project (TIN2017-85160-C2-1-R) from MINECO, and from Xunta de Galicia (ED431B 2017/01, ED431G 2019/01).

\bibliography{main,anthology}
\bibliographystyle{acl_natbib}

\end{document}